\documentclass[runningheads]{llncs}
\usepackage[utf8]{inputenc}
\usepackage[T1]{fontenc}
\usepackage[english]{babel}
\usepackage{subcaption}
\usepackage{url}
\usepackage{siunitx}
\usepackage{hyperref}

\usepackage{tikz}
\usetikzlibrary{arrows.meta, positioning, shapes.geometric}

\tikzset{
block/.style = {rectangle, draw, rounded corners,
                minimum height=0.9cm, minimum width=3cm,
                align=center},
decision/.style = {diamond, draw, aspect=2,
                   align=center, inner sep=1pt},
data/.style = {trapezium, draw,
               trapezium left angle=70,
               trapezium right angle=110,
               align=center},
line/.style = {draw, -{Latex[length=3mm]}, thick}
}
\usepackage{pgfplots}
\pgfplotsset{compat=1.18}
\usepgfplotslibrary{colorbrewer}
\usepackage{amsmath}
\usepackage{amssymb}
\usepackage{amsfonts}
\usepackage[capitalise, noabbrev]{cleveref}
\usepackage{graphicx}
\usepackage{bbding}
\usepackage{cite}
\usepackage{lipsum}
\usepackage{microtype}

\usepackage{csquotes}

\usepackage{booktabs}

\usepackage{tikz}
\usepackage[edges]{forest}
\usepackage{wheelchart}
\usepackage{flowchart}
\usetikzlibrary{
    calc,
    backgrounds,
    bbox,
    decorations.text,
    fit,
    shapes.geometric,
    arrows,
    arrows.meta,
    angles,
    math,
    positioning,
    quotes
}

\usepackage{todonotes}
\setuptodonotes{inline}

\usepackage{siunitx}
\DeclareSIUnit\knot{kn}
\DeclareSIUnit\nauticalmile{NM}
\DeclareSIUnit\foot{ft}
\DeclareSIUnit\flops{FLOPS}
\DeclareSIUnit\pixel{px}
\DeclareSIUnit\point{pt}

\begin{document}
\title{From High-Dimensional Spaces to Verifiable ODD Coverage for Safety-Critical AI-Based Systems}
\titlerunning{Verifiable ODD Coverage in AI-Based Systems}
\author{
Thomas Stefani\orcidID{0000-0001-7352-0590}\textsuperscript{(\Envelope)}
%Thomas Stefani\orcidID{0000-0001-7352-0590}
\and
Johann Maximilian Christensen\orcidID{0000-0001-9871-122X}
\and
Elena Hoemann\orcidID{0000-0001-9315-548X}
\and
Frank Köster
\and
Sven Hallerbach
}
\authorrunning{T. Stefani et al.}
\institute{
Institute for AI Safety and Security, German Aerospace Center (DLR),\\
Sankt Augustin and Ulm, Germany\\
Corresponding Author: \email{thomas.stefani@dlr.de}
}
\maketitle              % typeset the header of the contribution
\begin{abstract}
While Artificial Intelligence (AI) offers transformative potential for operational performance, its deployment in safety-critical domains such as aviation requires strict adherence to rigorous certification standards.
Current EASA guidelines mandate demonstrating complete coverage of the AI/ML constituent's Operational Design Domain (ODD)---a requirement that demands proof that no critical gaps exist within defined operational boundaries.
However, as systems operate within high-dimensional parameter spaces, existing methods struggle to provide the scalability and formal grounding necessary to satisfy the completeness criterion, and no standardized engineering method exists to bridge the gap between abstract ODD definitions and verifiable evidence.
This paper addresses that gap with a structured, multi-step method that integrates parameter grouping, a dependency analysis distinguishing deterministic from stochastic relationships, physically grounded constraint definition, and criticality-based discretization into a single ODD coverage verification process.
Applied to a case study on AI-based near mid-air collision avoidance with the VerticalCAS system, the method reduces the verification space by roughly \qty{60}{\percent} while preserving all safety-relevant combinations and yields an explicit, iterable list of uncovered regions, providing a repeatable process for demonstrating the completeness EASA requires.
Designed to scale toward higher-dimensional ODDs, the method advances a Safety-by-Design approach to the certification of future AI-based flight systems.
\end{abstract}
\section{Introduction}
The integration of Artificial Intelligence (AI) and Machine Learning (ML) into safety-critical applications in aviation represents a transformative shift in aerospace engineering~\cite{PrecedenceResearch2022}.
While neural networks offer unprecedented capabilities for processing complex sensor data and optimizing real-time decision-making, such as in the next-generation Airborne Collision Avoidance Systems (ACAS X), their \emph{black-box} nature directly conflicts with traditional deterministic safety assurance~\cite{Julian2019a, Christensen2024, Christensen2025}.
In this highly regulated domain, the transition from \emph{experimental} to \emph{certified} systems depends not on performance alone, but on verifiable trust.
The European Union Aviation Safety Agency (EASA) has addressed this challenge by outlining a roadmap for AI trustworthiness~\cite{EUASA2023}, emphasizing the necessity of Operational Design Domain (ODD) coverage, since ODDs define the area where the AI-based system is intended to operate.
Current EASA guidelines mandate that for an AI/ML constituent to be certified, the applicant must demonstrate \enquote{completeness} of the ODD.
This implies proving that the system has been verified across all relevant operating conditions it might encounter, ensuring no safety gaps exist within the defined boundaries~\cite{EUASA2024}.
However, a fundamental engineering gap exists between the abstract definition of an ODD, which often comprises a mere list, and the practical demonstration of its coverage.
Safety-critical systems typically operate in high-dimensional parameter spaces, where variables such as relative velocity, closure rate, bearing, and environmental factors interact.
As dimensionality increases, the state space expands exponentially.
Traditional sampling or brute-force simulation cannot provide the formal completeness guarantees required by certification authorities, as they either leave vast regions of the high-dimensional space unexplored or prove to be computationally infeasible~\cite{Spaeth2024}.
Furthermore, current engineering frameworks lack a standardized method to bridge the gap between an abstract ODD definition and a demonstration of coverage.
While formal methods provide exactness, they often struggle to meet the scalability requirements of deep neural networks.
Conversely, simulation-based testing offers scalability but often at the cost of high computational effort needed to claim \emph{complete coverage} that satisfies certification authorities.
There is a pressing need for a systematic approach to identify safety-critical regions within the ODD and focus verification efforts there.

This paper addresses this challenge by proposing a structured method for verifiable ODD coverage that translates abstract ODD definitions into quantifiable, verifiable metrics.
To ensure computational viability in high-dimensional spaces, we introduce a specialized method for criticality-based dimension reduction.
Later, the method is validated through an empirical study of AI-based near mid-air collision (NMAC) avoidance, providing a blueprint for the rigorous evidence required to satisfy EASA's stringent ODD completeness standards.
By bridging high-dimensional data with regulatory compliance, this work advances a method that can be integrated into a Safety-by-Design methodology for the certification of future AI-based flight systems.

The remainder of this paper is organized as follows: The \cref{sec:Related_Work} reviews the regulatory context and existing approaches for parameter space coverage.
Next, \cref{sec:Methodolgy}  details the proposed multi-step method for demonstrating ODD completeness through discretization and constraint definition.
In \cref{sec:UseCase}, this method is applied to a case study on the VerticalCAS airborne collision avoidance system.
Finally, \cref{sec:Discussion}  discusses the engineering implications and scalability of the approach, followed by a conclusion and future outlook in \cref{sec:Conclusion}.

\section{Related Work}\label{sec:Related_Work}
The safety assurance of AI-based systems in aviation is situated at the intersection of upcoming regulatory requirements and the novel engineering challenges of demonstrating compliance for AI-based systems.
This section reviews the current state of the art in both the regulatory context and ODD coverage to contextualize the proposed method.

\subsection{Regulatory Context}
The EASA Concept Paper, \emph{Guidance for Level 1 \& 2 Machine Learning Applications}, establishes a framework for AI trustworthiness in aviation, centering on Learning Assurance~\cite{EUASA2024}.
A critical requirement within this framework is the rigorous definition and verification of the ODD, as detailed in multiple objectives, including DM-08 and LM-16.
These guidelines mandate that applicants demonstrate not only the representativeness of data but also the completeness of the verification process across all operational dimensions.
EASA specifically requires evidence that no safety gaps exist within the ODD, suggesting quantitative approaches, such as subdividing the parameter space into hypercubes, to assess data density and model stability.
However, as the ODD encompasses increasingly complex, high-dimensional spaces, providing the evidence of completeness demanded by these objectives becomes a significant engineering hurdle that this paper addresses.

\subsection{Parameter Space Coverage}
Quantifying coverage in high-dimensional parameter spaces has become a fundamental requirement for evaluating the verification and operational safety of AI-based systems.
Weissensteiner~\cite{Weissensteiner2023} established a high-level ODD coverage process for automated driving systems, using an adaptive k-means clustering algorithm under predefined boundary conditions to reduce the number of required scenarios without compromising verification quality.
While this provides essential evidence for safety argumentation, the method does not account for complex interactions between parameters and therefore remains vulnerable to the \emph{curse of dimensionality}.

A complementary line of work addresses parameter dependence rather than scenario count. Höhndorf et al.~\cite{Hoehndorf2024} characterize dependence structures among operational parameters using vine copulas with subset simulation, optimizing failure-probability estimation by pruning implausible or low-likelihood combinations, and focusing verification on realistic regions.
Copulas, however, do not reduce the nominal dimensionality of the ODD: they impose a dependence structure on the same space, and each pairwise copula adds correlation parameters that may require non-trivial measurement.
Their cost thus grows with dimensionality, leaving open the need for a framework that scales to real-world ODDs while satisfying EASA's exhaustive completeness criteria.

A third group of approaches defines coverage geometrically.
Hirschle et al.~\cite{Hirschle2024} evaluate methods such as the convex hull of the sampled data, which gives a clear geometric boundary but is insensitive to internal density and scales poorly, and the union of hyperspheres around samples, which is highly sensitive to the chosen radius and can overestimate coverage beyond unity.
They propose kernel density estimation as a more scalable alternative for identifying low-density regions.

The idea of subdividing an ODD to make coverage tractable has recently been pursued at the ODD level by Schäfer and Cichon~\cite{schäfer2025incrementalvalidationautomateddriving}, who partition an ODD into micro-ODDs (µODDs) with generic-volume object representations, building a bottom-up completeness argument from manageable sub-domains rather than enumerating explicit instances.
The allocation of verification effort across such partitions has a longer history in software reliability engineering, where two limitations of a structural criterion are well established.
First, uniform coverage may misallocate effort: Cotroneo et al.~\cite{RELAI} instead sample inputs according to an operational profile in their RELAI scheme, contrasting with the strict, likelihood-agnostic completeness this work pursues.
Second, exercising a partition does not guarantee that faults within it surface, as captured by Voas's propagation–infection–execution (PIE) model~\cite{Voas}, in which fault detection requires not only that a location be executed but also that the execution infect the program state and that this state propagate to an observable output.
Covering a bin, in our sense, satisfies only the first condition—underscoring that coverage is a structural prerequisite for, rather than a guarantee of, correct behavior.

On the regulatory side, the ForMuLA report by EASA and Collins Aerospace~\cite{ForMuLA} proposes the closest analog to a completeness check: dividing the ML constituent ODD into subspaces at a required granularity and verifying that a sufficient number of examples populate each subspace, supported by Design-of-Experiment and Latin-hypercube sampling.
In their case study, however, this check is demonstrated only on a low-dimensional categorical cross-product (mission pattern against operating environment) and is assessed by visual inspection of the resulting distribution rather than by a quantitative metric.
The subspace granularity is left unspecified, and the sampling-based nature of the approach provides coverage in expectation rather than the exhaustive guarantee implied by EASA's completeness wording.

Thus, despite these advances in quantifying ODD sub-spaces, a distinct void remains: there is no systematic engineering process that bridges these individual metrics with EASA's overarching demand for ODD completeness, a prerequisite for the formal certification of AI-based systems~\cite{EUASA2024}.
This work addresses that void by turning the subspace idea into a quantitative, exhaustive completeness criterion over the discretized parameter space, equipped with methodical dimension reduction to keep it computationally feasible.

\section{Method for Demonstrating Completeness of the ODD}\label{sec:Methodolgy}
\begin{figure}[htbp]
\centering
\begin{tikzpicture}[node distance=1.6cm]
% Nodes
\node (start) [block] {Start};
\node (groupdec) [decision, below of=start, yshift=-0.3cm] {Grouping\\possible?};
\node (odd) [data, left=2cm of groupdec] {ODD};
\node (group) [block, right=1.8cm of groupdec] {Parameter\\Grouping};
\node (dep) [block, below of=groupdec, yshift=-0.3cm] {Dependency Analysis};
\node (constraints) [block, below of=dep] {Constraints Definition\\(deterministic / probabilistic)};
\node (bin) [block, below of=constraints] {Discretization / Binning};
\node (coverage) [block, below of=bin] {Coverage Testing for\\ \emph{Completeness}};
\node (exec) [data, right=2cm of coverage] {Executed\\Scenarios};
\node (covdec) [decision, below of=coverage] {Coverage\\Sufficient?};
\node (newscen) [block, right=1.6cm of covdec] {New Scenario\\Parameters};
\node (end) [block, below=1cm of covdec] {End};
% Connections
\path [line] (start) -- (groupdec);
\path [line] (odd) -- (groupdec);
\path [line] (groupdec) -- node[above]{Yes} (group);
\path [line] (groupdec) -- node[right]{No} (dep);
\path [line] (group) |- (dep);
\path [line] (dep) -- (constraints);
\path [line] (constraints) -- (bin);
\path [line] (bin) -- (coverage);
\path [line] (exec) -- (coverage);
\path [line] (coverage) -- (covdec);
\path [line] (covdec) -- node[right,yshift=-3mm]{No} (newscen);
\path [line] (newscen) -- (exec);
\path [line] (covdec) -- node[left]{Yes} (end);
\end{tikzpicture}
\caption{Process flow to comply with the coverage requirement of completeness by reducing high-dimensional spaces to a verifiable ODD contributing to a Safety-by-Design approach for AI-based Systems.}
\label{fig:process_flow2}
\end{figure}
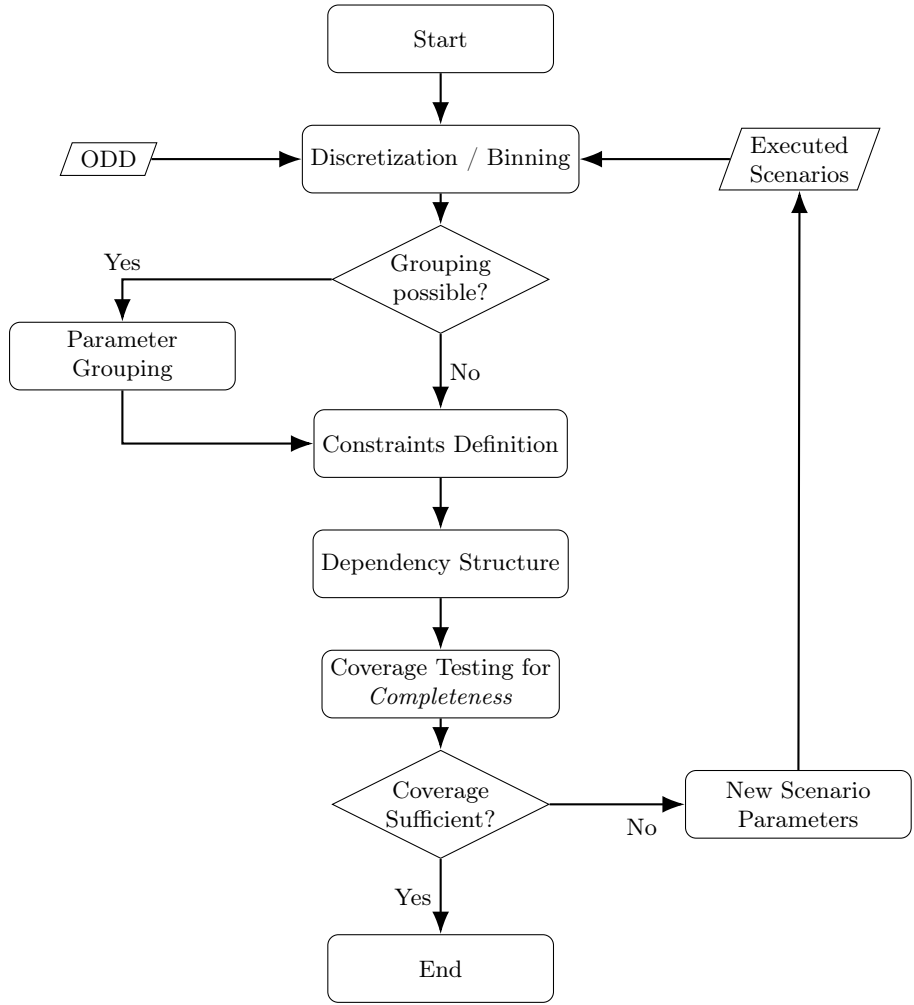

In EASA's Concept paper for AI learning assurance, the anticipated Means of Compliance (MOC) DM-08 requires \emph{\enquote{\dots coverage of the whole ODD by the test data set, with the necessary level of completeness
and representativeness.}}~\cite{EUASA2024}.
However, the process of demonstrating the necessary level of completeness remains undetermined.
This work provides a comprehensive method that engineers can follow.
It is part of a domain-agnostic effort to establish a Safety-by-Design AI engineering pipeline in which demonstrating adequate coverage is one component of the safety argumentation~\cite{Hoemann2026}.
\cref{fig:process_flow2} illustrates the workflow for assessing coverage completeness in a set of scenarios.
It begins with the defined ODD, which serves as input to the first step of parameter grouping.
Subsequently, the reduced parameter space feeds into the dependency analysis, constraint definition, and discretization steps before coverage testing is performed against the database of executed scenarios.
A decision step first checks whether \emph{Parameter Grouping} is possible.
If possible, parameters may be grouped to reduce the dimensionality of the coverage space before any further analysis is performed.
For example, multiple ODD parameters describing similar conditions, such as precipitation type and precipitation intensity, could be combined into a single precipitation condition parameter.
Such grouping reduces the number of parameter combinations to consider during coverage testing.
However, it is important to ensure that the grouping does not mask safety-relevant distinctions and therefore does not reduce the effectiveness of the coverage analysis.
To this end, grouped parameters should be validated, e.g., through sensitivity analysis or by temporarily refining the grouped ranges, to confirm that no additional safety-critical behaviors appear.
This step operates at the schema level, and should therefore precede all subsequent steps, since grouping changes which parameters exist and thus which dependencies, constraints, and bin sizes need to be defined.

In the \emph{Dependency Analysis} step, the relationships between the remaining parameters are characterized before any space reduction is performed.
This is a prerequisite for the subsequent constraint definition step, as the nature of the dependency---whether deterministic or stochastic---determines the appropriate reduction method.
Deterministic dependencies arise when the relationship between parameters can be derived analytically from physical laws or system kinematics.
In such cases, the dependency structure is fully known and can be expressed as an exact analytical constraint.
Stochastic dependencies, by contrast, arise when no closed-form relationship exists and the joint parameter distribution must be estimated from data.
In these cases, statistical methods such as vine copulas, Bayesian networks, or kernel density estimation may be employed to characterize the joint probability distribution of ODD parameters~\cite{Hoehndorf2024}.
When relying on such estimated dependencies, it is essential to distinguish correlation from causation, as a statistical association between parameters does not necessarily reflect a causal relationship and may yield misleading constraints if treated as one.
It is important to note, however, that such statistical methods do not reduce the dimensionality of the ODD itself; they impose a joint probability distribution over the full parameter space rather than eliminating dimensions.
Consequently, if the completeness criterion demands that all bin combinations be covered regardless of their likelihood, statistical dependency models provide no direct benefit to the coverage metric.
Their utility lies instead in informing the subsequent constraint definition step, where low-likelihood regions can be filtered, or in prioritizing scenario generation toward high-likelihood uncovered regions.

In the \emph{Constraints Definition} step, the insights from the dependency analysis are translated into concrete reductions of the parameter space.
For deterministic dependencies, analytical constraint functions are derived directly from the known physical relationships, removing provably unreachable or irrelevant bin combinations from the required coverage set.
For stochastic dependencies, likelihood-based filters derived from the estimated joint distribution can be used to exclude parameter combinations with negligible probability of occurrence, provided this probabilistic relaxation of the completeness criterion is accepted by the relevant certification authority.
Beyond dependency-driven constraints, this step also identifies combinations that are physically implausible, operationally irrelevant, or of lower safety criticality.
Such combinations can be excluded or tested more sparsely without significantly affecting the assessment of safety-relevant coverage.
The result of this step is a reduced parameter space containing only those combinations that are both physically plausible and safety-relevant.

In the \emph{Discretization and Binning} step, the reduced parameter space is divided into a finite number of bins per parameter, since continuous parameters make exhaustive coverage of the parameter space infeasible.
Performing discretization after grouping, dependency analysis, and constraint definition ensures that binning operates on the already-reduced relevant parameter space rather than the full ODD, avoiding unnecessary computational overhead.
The bin size is determined by the criticality $c$ of each parameter, which must be defined in advance and can be specified either uniformly or as a function of $c$.
This step is crucial, as it directly determines the number of bins and the corresponding bin combinations to be evaluated during coverage analysis.
If bins are chosen too large, relevant regions of the parameter space may remain insufficiently explored, potentially leaving safety-critical gaps undetected.
Conversely, selecting overly small bins significantly increases the number of combinations and can make the analysis computationally expensive.
Therefore, defining an appropriate criticality measure and deriving suitable bin sizes from it is an important step in balancing computational feasibility with sufficient resolution of the parameter space.
The method applies to ODD parameters that are either continuous or can be meaningfully mapped to a continuous or ordinal representation, such as translating 'time of day' categories into illuminance ranges.
Parameters that are purely categorical without such a mapping are outside the current scope and require separate treatment.

In the \emph{Coverage Testing} step, the discretized and reduced parameter space is systematically analyzed to determine whether the executed scenarios sufficiently cover the ODD.
Each parameter is divided into a finite number of bins, and coverage is evaluated across all combinations of these bins.
Let $B_i$ denote the set of bins for parameter $p_i$ after grouping and discretization.
The full discretized parameter space is given by the Cartesian product
\begin{align}
    \mathcal{B} = B_1 \times B_2 \times \dots \times B_n\text{.}
\end{align}
The preceding grouping, dependency analysis, and constraint-definition steps jointly yield a set of admissibility conditions $\mathcal{C}$.
Applying these conditions reduces the full space to the \emph{reduced} parameter space
\begin{align}
    \mathcal{B}_\mathrm{reduced} = \{ b \in \mathcal{B} \mid b \text{ satisfies all conditions in } \mathcal{C} \} \subseteq \mathcal{B}\text{.}
\end{align}
The reduction is intentionally conservative: a bin combination is removed only if it is provably infeasible or non-critical, while combinations of uncertain relevance are retained.
As a result, $\mathcal{B}_\mathrm{reduced}$ is an over-approximation of the truly relevant space and may still contain less-relevant bins, but no safety-relevant combination is discarded.
A bin combination $b \in \mathcal{B}$ is considered \emph{covered} if at least one data point from an executed scenario lies within the corresponding parameter ranges of $b$.
Coverage $r_\mathrm{cov}$ is considered sufficient only if all bin combinations in the reduced parameter space are covered, i.e.,
\begin{align}
    r_\mathrm{cov} = \frac{|\mathcal{B}_\mathrm{reduced, covered}|}{|\mathcal{B}_\mathrm{reduced}|} = 1\text{.}
\end{align}

If uncovered regions remain, the analysis produces a list of missing bin combinations.
These uncovered combinations can be translated into new scenario parameters, which are then executed and added to the existing scenario set.
The coverage analysis is then repeated, forming an iterative process that continues until the required coverage of the reduced parameter space is achieved.

\section{Case Study: AI-Based Collision Avoidance System}\label{sec:UseCase}

The Vertical Collision Avoidance System (VerticalCAS) is an advanced aircraft collision avoidance system currently under development.
It is designed as a successor to the widely deployed \emph{Traffic Collision Avoidance System II}, with the goal of improving vertical separation advisories and enhancing overall airborne safety.
VerticalCAS provides automated guidance to pilots by issuing resolution advisories based on the relative vertical motion and position of nearby aircraft, taking into account updated sensing, prediction, and advisory logic compared to its predecessor~\cite{Kochenderfer2012}.
The corresponding advisories are listed in~\cite{Julian2019}.
\cref{tab:vertical_cas} lists the key state variables of the VerticalCAS system~\cite{Julian2019}.
It includes relative altitude ($h$), own and intruder vertical rates 
\((\dot{h}_\mathrm{own}, \dot{h}_\mathrm{int})\), the time to closest point of approach (CPA) (\(\tau\)), and the previous advisory indicator ($s_\mathrm{adv}$), along with their corresponding ranges of values.
These variables define the state space and are used as the ODD for VerticalCAS.
The values are oriented on Kochenderfer et al.~\cite{Julian2019} but differ sligtly since in prior work~\cite{Christensen2024a, Christensen2024, Stefani2024a, Stefani2024b}, other scenario settings were used, and the main purpose of this work is to demonstrate the process flow rather than focus on the actual numbers.
\begin{table}[htb]
\centering
\caption{VerticalCAS State Variables and the parameter ranges, oriented on the GitHub release \url{https://github.com/sisl/VerticalCAS}}
\label{tab:vertical_cas}
\begin{tabular}{llcrr}
\toprule
\textbf{Variable} & \textbf{Description} & \textbf{Values} & \textbf{Bins} & \textbf{Bins adj.}\\
\midrule
\(h\) & Relative intruder altitude & $[\qty{-1500}{\foot}, \qty{1500}{\foot}]$ & 100 & 100 \\
\(\dot{h}_\mathrm{own}\) & Ownship vertical rate & $[\qty{-3200}{\foot\per\minute}, \qty{3200}{\foot\per\minute}]$ & 32 & 32 \\
\(\dot{h}_\mathrm{int}\)  & Intruder vertical rate & $[\qty{-3200}{\foot\per\minute}, \qty{3200}{\foot\per\minute}]$ & 32 & 1\\
\(\tau\)
& Time to CPA & $[\qty{0}{\second}, \qty{60}{\second}]$ & 60 & 60 \\
\(s_\mathrm{adv}\) & Previous advisory & see~\cite{Julian2019} & 9 & 1 \\
\bottomrule
\end{tabular}
\end{table}
Data from our prior work was used to populate the database for the scenarios executed.
For this, the data was aggregated and stored in a unified CSV file with 1.97 million rows of VerticalCAS state variables.

\paragraph{Parameter Grouping.}
In this work, parameter grouping is not applied.
The VerticalCAS state space under consideration already comprises a small set of well-defined parameters, making additional grouping unnecessary.
However, this step is intended for larger ODDs with many parameters, where multiple related variables can be combined into higher-level parameters, reducing the dimensionality of the scenario space and the number of bin combinations to be analyzed.
\paragraph{Dependency Analysis.}
Before any reduction is performed, the relationships among the VerticalCAS state variables are characterized to determine their nature.
Two dependencies dominate the relevant encounter dynamics.
First, the safety-relevant range of the relative altitude $h$ depends on the time to CPA $\tau$: for large $\tau$, the full altitude range can still evolve into a conflict, whereas for small $\tau$ only small relative altitudes can.
Second, the criticality of an encounter depends on the joint configuration of $h$ and the relative vertical rate $\dot{h}_\mathrm{rel}$, since their relative sign determines whether the aircraft are converging or diverging.
Both relationships follow directly from the vertical kinematics of the encounter and are therefore \emph{deterministic}: they admit a closed-form description rather than requiring estimation from data, and are accordingly handled in the next step through exact analytical constraints.
Stochastic dependency models such as vine copulas, Bayesian networks, or kernel density estimation, by contrast, are designed to characterize dependence structures that have no closed-form description and must be inferred from observed data~\cite{Hoehndorf2024}.
As the governing dependencies in this use case are deterministic and the parameter space is low-dimensional, there is no stochastic dependence structure to estimate, so such methods would add modeling cost without refining the reduction.
They are therefore not applied here, but remain the appropriate tool for larger ODDs in which parameters exhibit genuine stochastic interactions.

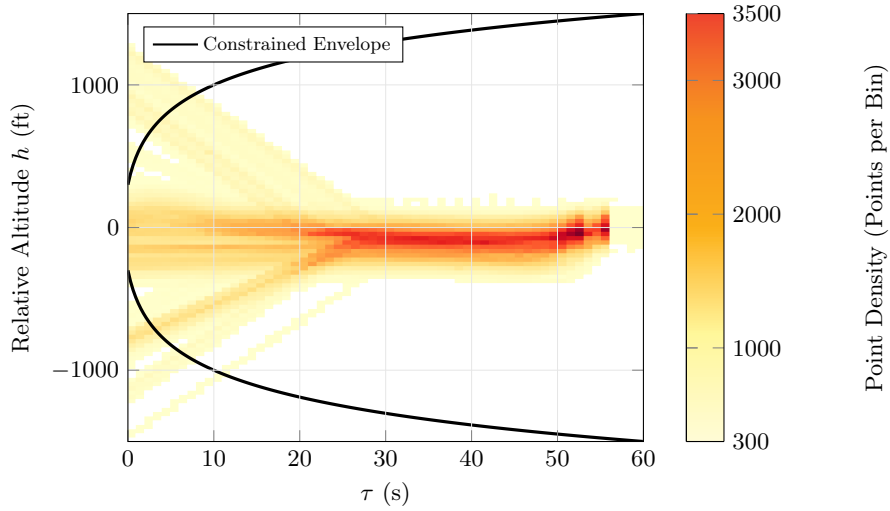
\begin{figure}[htbp]
\centering
\begin{tikzpicture}
    \begin{axis}[
        axis on top,
        title={},
        xlabel={\(\tau\) (s)},
        ylabel={Relative Altitude \(h\) (ft)},
        yticklabel style={/pgf/number format/1000 sep={}        },
        xmin=0, xmax=60,
        ymin=-1500, ymax=1500,
        grid=both,
        legend image post style={black},
        legend style={nodes={scale=0.8, transform shape}},
        legend image code/.code={
        \draw[black, thick] (0cm,0cm) -- (0.6cm,0cm);
        }
        grid style={line width=.1pt, draw=gray!10},
        major grid style={line width=.2pt, draw=gray!20},
        legend pos=north west,
        colorbar,
        point meta min=300,
        point meta max=3500,
        colormap={heatmap}{
            color(0cm)=(yellow!20);
            color(1.25cm)=(yellow!50);
            color(2.5cm)=(orange!70!yellow);
            color(3.75cm)=(orange);
            color(5cm)=(red!90)
        },
        colorbar style={
            ylabel={Point Density (Points per Bin)},
            ylabel style={
                font=\small,
                at={(4.5,0.5)},
                anchor=north
            },
            ytick={300, 1000, 2000, 3000, 3500},
            yticklabel style={font=\footnotesize, /pgf/number format/1000 sep={}}
        }
    ]
    % 1. HEATMAP IMAGE OVERLAY
    \addplot graphics [xmin=0, xmax=60, ymin=-1500, ymax=1500] {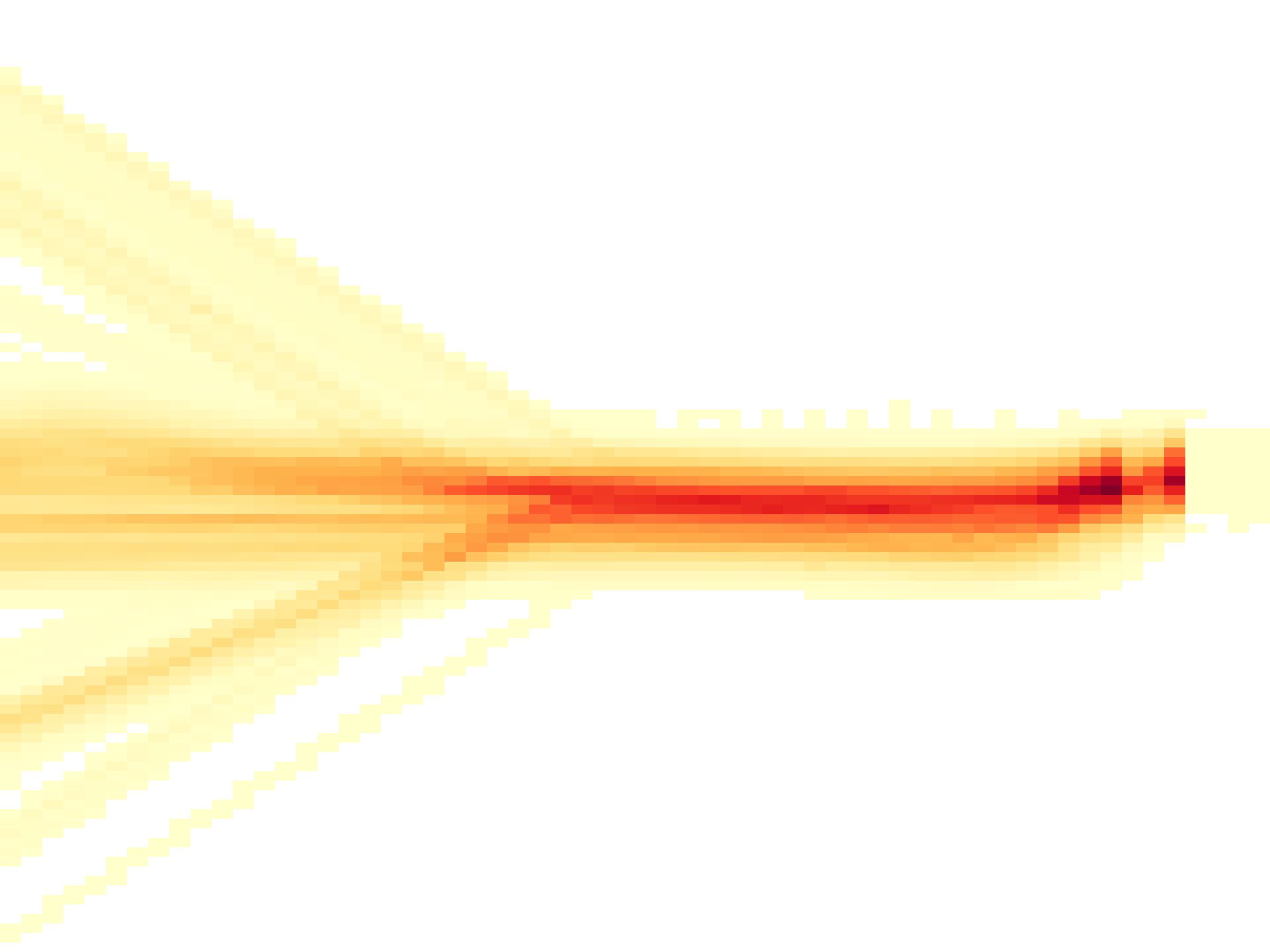};

    % 2. Upper Constrained Envelope: min(1500, 100 + (3200/60)*tau)
    \addplot [
        domain=0:60,
        samples=400,
        black,
        very thick,
        forget plot
    ] {min(1500, 100 + (3200/60)*x)};

    % 3. Lower Constrained Envelope
    \addplot [
        domain=0:60,
        samples=400,
        black,
        very thick
    ] {-min(1500, 100 + (3200/60)*x)};

    \addlegendentry{Constrained Envelope}
    \end{axis}
\end{tikzpicture}
\caption{Example visualization of $h_\mathrm{rel}$ over $\tau$. The heatmap indicates the number of points contained in each bin in the dataset. The black line indicates the applied constrained envelope.}
\label{fig:plot}
\end{figure}

\paragraph{Constraint Definition.}
The two deterministic dependencies identified above are now translated into explicit admissibility conditions that remove provably infeasible or non-critical bin combinations.
The first constraint operationalizes the $h$--$\tau$ coupling.
The bound is derived directly from the vertical kinematics of the encounter.
A conflict at the CPA requires the relative altitude to be reduced below the NMAC threshold
$h_\mathrm{NMAC}$ within the remaining time $\tau$.
In the considered scenarios, the intruder maintains level flight, so the relative vertical motion is governed by the ownship alone, whose vertical rate is limited to \qty{\pm3200}{\foot\per\minute}, see \cref{tab:vertical_cas}.
The relative vertical closing rate is therefore bounded in the worst case by $\dot{h}_\mathrm{rel, max} = \qty{3200}{\foot\per\minute}$, attained when the gap is closed at the maximum rate; for encounters in which the intruder may also maneuver vertically, the same construction applies with $\dot{h}_\mathrm{rel, max}$ given by the sum of both aircraft's maximum rates. The largest relative altitude from which co-altitude is still reachable within $\tau$ is therefore
\begin{align}
    h_\mathrm{max}(\tau) = \min\left(h_\mathrm{ODD}, h_\mathrm{NMAC} + \dot{h}_\mathrm{rel,max}\,\tau\right)\text{,}
\end{align}
with $h_\mathrm{ODD} = \qty{1500}{\foot}$ and $h_\mathrm{NMAC} = \qty{100}{\foot}$.
States above this bound are highly unlikely to reach a conflict within the available time and are excluded from the coverage analysis. Because the exclusion rests on kinematic reachability rather than on an assumed likelihood, no state capable of producing a conflict is discarded.
The bound grows linearly with $\tau$ and reaches the full ODD range at $\tau = (h_\mathrm{ODD} - h_\mathrm{NMAC})/\dot{h}_\mathrm{rel, max} \approx \qty{26}{\second}$, beyond which the entire altitude range remains safety-relevant.
It should be noted that deriving such constraints generally requires domain knowledge of the system's dynamics.
The resulting envelope is shown in~\cref{fig:plot}, where $h$ is plotted over $\tau$ with a heatmap of the per-bin datapoint count; as the underlying scenarios were not designed to demonstrate coverage, substantial uncovered area remains.
The second constraint operationalizes the geometric coupling between $h$ and $\dot{h}_\mathrm{own}$.
Encounters in which the aircraft are vertically diverging---ownship below the intruder ($h > 0$) while descending 
($\dot{h}_\mathrm{own} < 0$), or above the intruder ($h < 0$) while climbing ($\dot{h}_\mathrm{own} > 0$)---produce increasing vertical separation over time and are therefore non-critical.
These states are removed, retaining only converging or level-flight configurations for the coverage analysis.

\paragraph{Discretization and Binning.}
The bin size of each parameter is governed by its criticality $c$: the more a region of a parameter influences the safety of the advisory, the finer it must be resolved.
Here, criticality is assessed from expert knowledge, although a more detailed process could derive its distribution from simulation-based analyses.
Criticality may not be constant across a parameter range.
For $\tau$, for instance, an advisory issued with enough time to react is far more valuable than one issued a fraction of a second before a potential collision, which would suggest a finer resolution at reasonable larger $\tau$.
For the purpose of this demonstration, however, we assume a constant criticality distribution, which yields constant bin sizes for all parameters.
The resulting bin counts are listed in \cref{tab:vertical_cas}.
For $h$, $\dot{h}_\mathrm{own}$, and $\dot{h}_\mathrm{int}$, slightly different bin sizes than in~\cite{Julian2019} are used, while for $\tau$ a resolution of one bin per second is retained.
The bin size for $h$ is set slightly below the height of an A320~\footnote{https://www.airbus.com/en/products-services/commercial-aircraft/passenger-aircraft/a320-family} to provide an adequate resolution for the advisory prediction.
Taken together, the full discretization yields \num{55296000} states.
This space is reduced by abstracting two variables that do not describe the current encounter geometry.
The state variable $s_\mathrm{adv}$ encodes the advisory issued in the previous time step and originally takes nine values.
Its role is to maintain advisory consistency and to prevent oscillating advisories across time steps, rather than to characterize the geometric configuration.
As the coverage analysis here targets whether appropriate advisories are generated for the relevant encounter geometries, $s_\mathrm{adv}$ is abstracted to a single bin.
For the same reason, the intruder vertical rate $\dot{h}_\mathrm{int}$ is held fixed in the considered scenarios and is likewise represented by a single bin, so that the relative vertical motion is captured through $\dot{h}_\mathrm{own}$.
Together, these two abstractions reduce the parameter space from \num{55296000} by \qty{99.7}{\percent} to \num{192000} unconstrained states.

\paragraph{Coverage Testing for Completeness.}

\cref{tab:coverage_report_sidebyside} summarizes the coverage metrics for the discretized VerticalCAS state space.
The total unconstrained state space contains \num{192000} combinations, of which \num{6455} are observed in the dataset, yielding a raw coverage of \qty{3.36}{\percent}.
After applying the tailored constraints---including the $\tau$-dependent altitude envelope and removal of non-critical diverging encounters---the valid theoretical space is reduced by \qty{60.2}{\percent} to \num{76416} combinations.
Within this tailored space, \num{2001} combinations are observed, yielding a tailored coverage of \qty{2.62}{\percent}.
While the constrained coverage is lower in this example, it demonstrates that the effort required to achieve full coverage with newly generated scenarios is reduced because only physically meaningful and relevant combinations are left to consider.
\begin{table}[htb]
\centering
\caption{Coverage report for VerticalCAS: Unconstrained vs.\ Constrained}
\label{tab:coverage_report_sidebyside}
\begin{tabular}{lrr}
\toprule
\textbf{Metric} & \textbf{Unconstrained} & \textbf{Constrained} \\
\midrule
Combinations & \num{192000} & \num{76416} \\
Combinations Covered & \num{6455} & \num{2001} \\
Coverage & \qty{3.36}{\percent} & \qty{2.62}{\percent} \\
\bottomrule
\end{tabular}
\end{table}%

In addition, the script used for this coverage analysis generates a list of missing combinations in the theoretical state space.
These missing combinations correspond to regions of the parameter space that have not yet been observed in the dataset.
By translating these combinations back into physical values, new scenarios can be created and added to the dataset.
Iteratively generating and executing such scenarios allows coverage to be systematically increased, ultimately achieving \emph{complete} coverage of the constrained operational envelope.

% \begin{table}[t]
%   \centering
%   \caption{Sample of valid but uncovered (missing) state-space bins under the tailored envelope. $\dot{h}_\mathrm{int}$ and $s_\mathrm{adv}$ are single-bin and omitted.}
%   \label{tab:missing-tailored}
%   \begin{tabular}{rrr}
%     \toprule
%     $h$ [ft] & $\dot{h}_\mathrm{own}$ [ft/min] & $\tau$ [s] \\
%     \midrule
%     $[-1500, -1470)$ & $[-3200, -3000)$ & $[26, 27)$ \\
%     $[-1500, -1470)$ & $[-3200, -3000)$ & $[27, 28)$ \\
%     $[-1500, -1470)$ & $[-3200, -3000)$ & $[28, 29)$ \\
%     $[-1500, -1470)$ & $[-3200, -3000)$ & $[29, 30)$ \\
%     $[-1500, -1470)$ & $[-3200, -3000)$ & $[30, 31)$ \\
%     $[-1500, -1470)$ & $[-3200, -3000)$ & $[31, 32)$ \\
%     $[-1500, -1470)$ & $[-3200, -3000)$ & $[32, 33)$ \\
%     $[-1500, -1470)$ & $[-3200, -3000)$ & $[33, 34)$ \\
%     \bottomrule
%   \end{tabular}
% \end{table}

\section{Discussion}\label{sec:Discussion}
The transition of AI into safety-critical aviation relies on verifiable trust.
Our method bridges abstract ODD definitions and EASA's completeness requirement, turning the loosely defined notion of covering \emph{the whole ODD} into a quantitative, exhaustive criterion over a discretized parameter space---a systematic, end-to-end process where none previously existed.
By evaluating
completeness across the Cartesian product of discretized bins rather than parameter-wise, it verifies that all parameter ranges are covered across all others, exposing safety gaps in interdependent regions of the ODD that independent parameter testing would overlook, even in high-dimensional state spaces.
Two design choices carry implications beyond the case study.
The first is how dependencies are treated: statistical models such as vine copulas impose a joint distribution but do not reduce dimensionality, so under a strict criterion requiring every bin combination regardless of likelihood, they offer no direct benefit to the coverage metric.
Their role is to inform constraint definition and prioritize scenario generation; genuine reduction instead stems from grouping, the abstraction of non-geometric variables, and constraints that provably remove unreachable or non-critical combinations.
The second is how those constraints are derived: grounding the $\tau$-dependent altitude bound directly in the encounter's vertical kinematics makes exclusion rest on physical reachability rather than on an assumed likelihood.
The reduction is therefore conservative---a combination is removed only if provably infeasible or non-critical---so the reduced space over-approximates the truly relevant one and discards no safety-relevant combination.
Performing binning last, on the already-reduced space, keeps the process tractable.
Applied to existing simulation data, the method yielded relatively low coverage in the VerticalCAS case study---expected, since the original scenarios were not designed for ODD completeness; the result's value lies in pinpointing untested operational regions and translating them into new scenario parameters.
Coverage here is an exercising criterion: a bin counts as covered once a single scenario falls within it, demonstrating that the region was tested, but not that the model behaves correctly there.
It is thus necessary but not sufficient for safety---a structural complement to behavioral verification rather than a substitute.
Two limitations frame the contribution.
First, coverage is defined over the ODD's input parameter space, not the ML constituent's learned latent representation.
Input-space completeness ensures the operational conditions an applicant must justify are systematically exercised, but does not guarantee uniform coverage of the internal feature space, where semantically distinct inputs may collapse to nearby representations or nearby inputs diverge.
The method thus targets EASA's operational completeness rather than AI-specific failure modes such as latent-space clustering or distribution-shift sensitivity; we regard parameter and latent-space coverage as complementary.
Second, weighting all bins equally can direct effort toward operationally improbable regions; the optional likelihood-based relaxation, filtering low-probability combinations subject to certification-authority acceptance, is the intended mechanism for this trade-off.

Finally, the VerticalCAS study was kept compact so the method's mechanics remain transparent; it is therefore a proof of concept rather than an operational-scale demonstration.
For high-dimensional ODDs, where the unconstrained space grows exponentially, the advanced stages---criticality-based bin sizing, parameter grouping, and stochastic dependency modeling for constraint definition---are what keep verification tractable while preserving safety-relevant coverage.
Demonstrating the method at that scale is the principal direction for future work.

\section{Conclusion and Outlook}\label{sec:Conclusion}
This paper addresses the engineering gap between abstract ODD definitions and the \emph{completeness} evidence required by EASA for AI/ML certification.
We proposed a structured, end-to-end method that reduces the parameter space through grouping, dependency analysis, and constraint definition before testing coverage exhaustively over the Cartesian product of the discretized bins.
A central element is constraint-based dimension reduction, which restricts verification to physically plausible and safety-relevant regions and thereby keeps the completeness check computationally feasible in high-dimensional spaces.
Applied to the VerticalCAS collision-avoidance use case, the method reduced the unconstrained state space by approximately \qty{60}{\percent} (from \num{192000} to \num{76416} combinations) while preserving all safety-relevant combinations, and the same analysis yields an explicit list of uncovered bins that can be translated into new scenario parameters.
This iterative gap-closing loop turns EASA's qualitative completeness requirement into a quantifiable, repeatable engineering procedure.
It contributes a building block toward a Safety-by-Design approach for future AI-based systems.
Several directions remain open. Because the case study deliberately used a compact parameter set, the parameter-grouping step---intended to combine related variables and curb combinatorial growth in larger ODDs---was not exercised, and demonstrating it on a higher-dimensional ODD with many interrelated parameters is a primary next step.
In the same setting, stochastic dependency modeling, e.g., through vine copulas, Bayesian networks, or kernel density estimation, can inform constraint definition where relationships are not analytically known.
Further work will replace the uniform criticality assumption with empirically or simulation-derived criticality distributions, yielding non-uniform bin resolutions that concentrate verification effort where it matters most.
Finally, extending the coverage criterion from the input parameter space toward the model's learned representation, and coupling it with behavioral assurance evidence, would broaden the method from operational completeness toward AI-specific safety guarantees.
As a concrete step in that direction, and because covering a bin only exercises it rather than verifying behavior, coupling the criterion with fault-injection or mutation-based analysis in the spirit of the PIE model would yield stronger per-region evidence.

\begin{credits}
% \subsubsection{\ackname} A bold run-in heading in small font size at the end of the paper is
% used for general acknowledgments, for example: This study was funded

\subsubsection{\discintname}
The authors have no competing interests to declare that are
relevant to the content of this article.
\end{credits}
\bibliographystyle{splncs04}
\bibliography{literature-bibtex}
%\label{bibliography}
\end{document}